\documentclass[letterpaper,journal]{IEEEtran}
\usepackage{amsmath,amsfonts}
\usepackage{algorithmic}
\usepackage{algorithm}
\usepackage{array}
\usepackage[caption=false,font=normalsize,labelfont=sf,textfont=sf]{subfig}
\usepackage{textcomp}
\usepackage{stfloats}
\usepackage{url}
\usepackage{verbatim}
\usepackage{graphicx}
\usepackage{bm}
\usepackage{float}
\def\BibTeX{{\rm B\kern-.05em{\sc i\kern-.025em b}\kern-.08em
    T\kern-.1667em\lower.7ex\hbox{E}\kern-.125emX}}
\usepackage{balance}
\usepackage[numbers,sort&compress]{natbib}
\usepackage[table]{xcolor}
\usepackage{booktabs}
\usepackage{multirow}
\usepackage[pagebackref=false,breaklinks=true,colorlinks,bookmarks=True]{hyperref}
\usepackage{CJKutf8}

\begin{document}
\begin{CJK}{UTF8}{gbsn}

\title{Multi-Text Guided Few-Shot Semantic Segmentation}

\author{Qiang Jiao, Bin Yan, Yi Yang, Mengrui Shi, Qiang Zhang
\thanks{The authors are with the State Key Laboratory of Electromechanical Integrated Manufacturing of High-Performance Electronic Equipments, and also with the Center for Complex Systems, School of Mechano-Electronic Engineering, Xidian University, Xi'an, Shaanxi 710071, China (e-mail: qjiao@xidian.edu.cn; [yanbin, yangyi\_iy, mrshi]@stu.xidian.edu.cn; qzhang@xidian.edu.cn).}
\thanks{Corresponding author: Qiang Zhang.}
}

\maketitle

\begin{abstract}
Recent CLIP-based few-shot semantic segmentation methods introduce class-level textual priors to assist segmentation by typically using a single prompt (e.g., ``a photo of \{class\}”). However, these approaches often result in incomplete activation of target regions, as a single textual description cannot fully capture the semantic diversity of complex categories. Moreover, they lack explicit cross-modal interaction and are vulnerable to noisy support features, further degrading visual prior quality.
To address these issues, we propose the Multi-Text Guided Few-Shot Semantic Segmentation Network (MTGNet), a dual-branch framework that enhances segmentation performance by fusing diverse textual prompts to refine textual priors and guide the cross-modal optimization of visual priors.
Specifically, we design a Multi-Textual Prior Refinement (MTPR) module that suppresses interference and aggregates complementary semantic cues to enhance foreground activation and expand semantic coverage for structurally complex objects.
We introduce a Text Anchor Feature Fusion (TAFF) module, which leverages multi-text embeddings as semantic anchors to facilitate the transfer of discriminative local prototypes from support images to query images, thereby improving semantic consistency and alleviating intra-class variations.
Furthermore, a Foreground Confidence-Weighted Attention (FCWA) module is presented to enhance visual prior robustness by leveraging internal self-similarity within support foreground features. It adaptively down-weights inconsistent regions and effectively suppresses interference in the query segmentation process.
Extensive experiments on standard FSS benchmarks validate the effectiveness of MTGNet. In the 1-shot setting, it achieves 76.8\% mIoU on PASCAL-$\bm{5^i}$ and 57.4\% on COCO-$\bm{20^i}$, with notable improvements in folds exhibiting high intra-class variations.
\end{abstract}

\begin{IEEEkeywords}
  Few-shot segmentation, Multi-text guided, Few-shot learning, CLIP
\end{IEEEkeywords}

\section{Introduction}
\IEEEPARstart{F}ew-shot semantic segmentation (FSS)~\cite{shaban2017one,wang2019panet} aims to segment unseen categories with only a few labeled support examples.
In FSS, support images often fail to encompass the full diversity of visual appearances within a given class. Moreover, the substantial intra-class variations between support and query images frequently hinder accurate segmentation in complex scenes. To mitigate these issues, existing methods have explored enhancing prototype representations~\cite{zhang2020sg, fan2022self, cao2023break, yang2020prototype, li2021adaptive} or refining matching mechanisms~\cite{hong2022cost, tian2020prior}. However, these methods remain limited when dealing with visually complex categories 
with high intra-class variations, as well as when the guidance information from support examples is sparse or poorly aligned.

\begin{figure}[t]
    \centering
    \includegraphics[width=\linewidth]{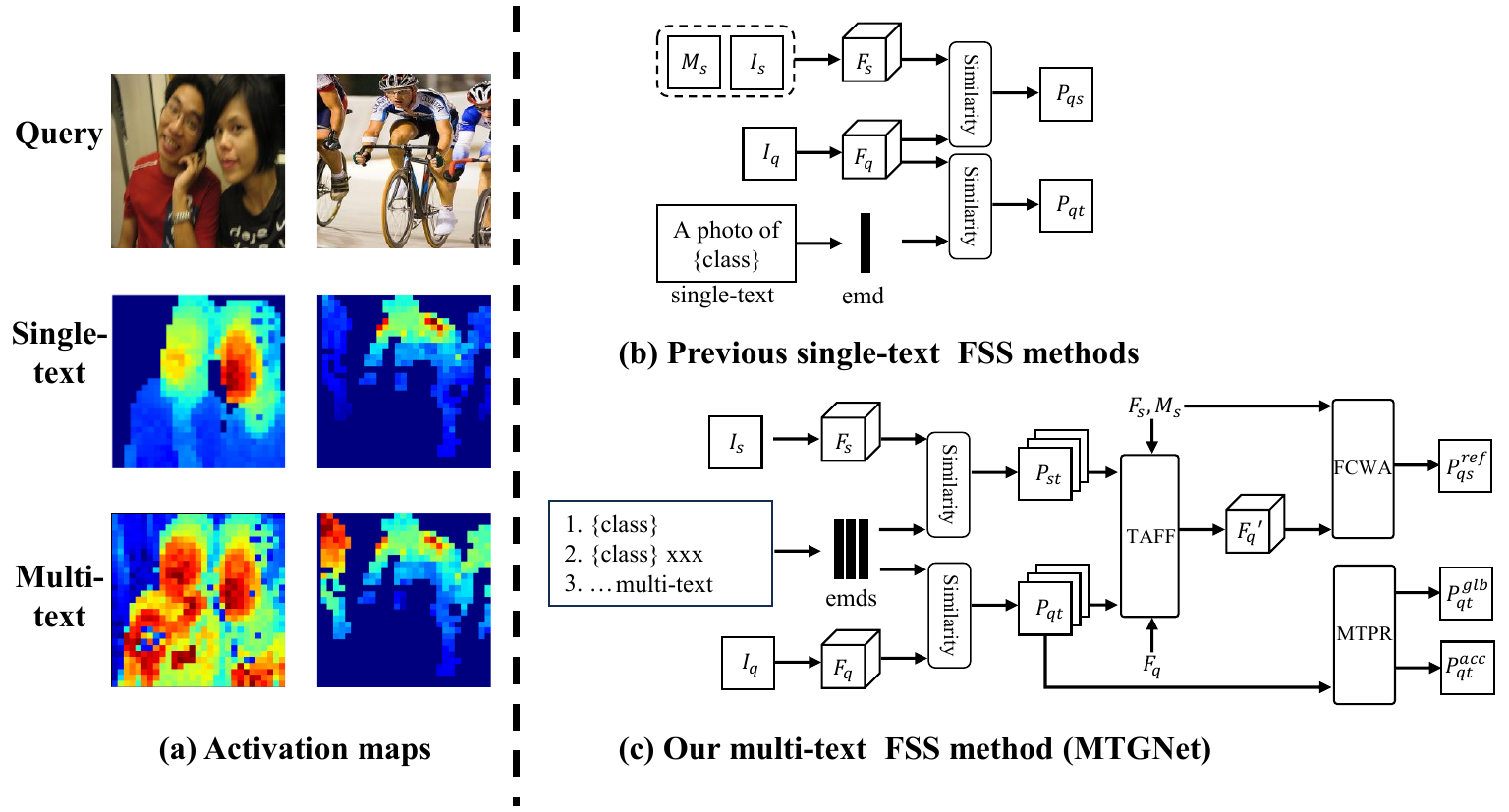}
    \caption{Comparisons of our MTGNet with previous single-text-based FSS methods. (a) Activation maps. Compared to single-text methods (middle row), which highlight only the most distinctive parts of the target object, our multi-text strategy (bottom row) activates a broader and more complete semantic region. (b) Previous single-text FSS methods. (c) Our proposed multi-text FSS method (MTGNet).}
    \label{fig:motivation}
    \vspace{-10pt}
  \end{figure}

In recent years, the Contrastive Language-Image Pretraining (CLIP) method~\cite{radford2021learning} has gained attention for its ability to align visual and textual modalities through large-scale image-text contrastive learning.
Inspired by this, several CLIP-based FSS approaches~\cite{luddecke2022image, leng2025multi, wang2024rethinking, chen2024visual} have introduced additional textual semantic priors to assist the segmentation process.
As illustrated in Fig.~\ref{fig:motivation}(b), these methods typically adopt a single prompt (e.g., ``a photo of \{class\}'') as an auxiliary input to generate a textual prior $P_{qt}$, which is subsequently combined with the visual prior $P_{qs}$ to guide the segmentation prediction. Although this strategy improves segmentation performance, it suffers from a critical limitation: 
\textbf{A single textual description is often inadequate to fully represent the semantic richness of complex categories, leading to incomplete foreground activation.} As shown in the second row of Fig.~\ref{fig:motivation}(a), when segmenting a compound class such as “person”, the resulting activation map from a single prompt captures only partial regions (e.g., the face or upper body), missing crucial contextual elements (e.g., clothing or lower limbs). Such incomplete activation easily results in substantial omission of relevant target pixels in the final segmentation.

\begin{figure}[t]
    \centering
    \includegraphics[width=0.9 \linewidth]{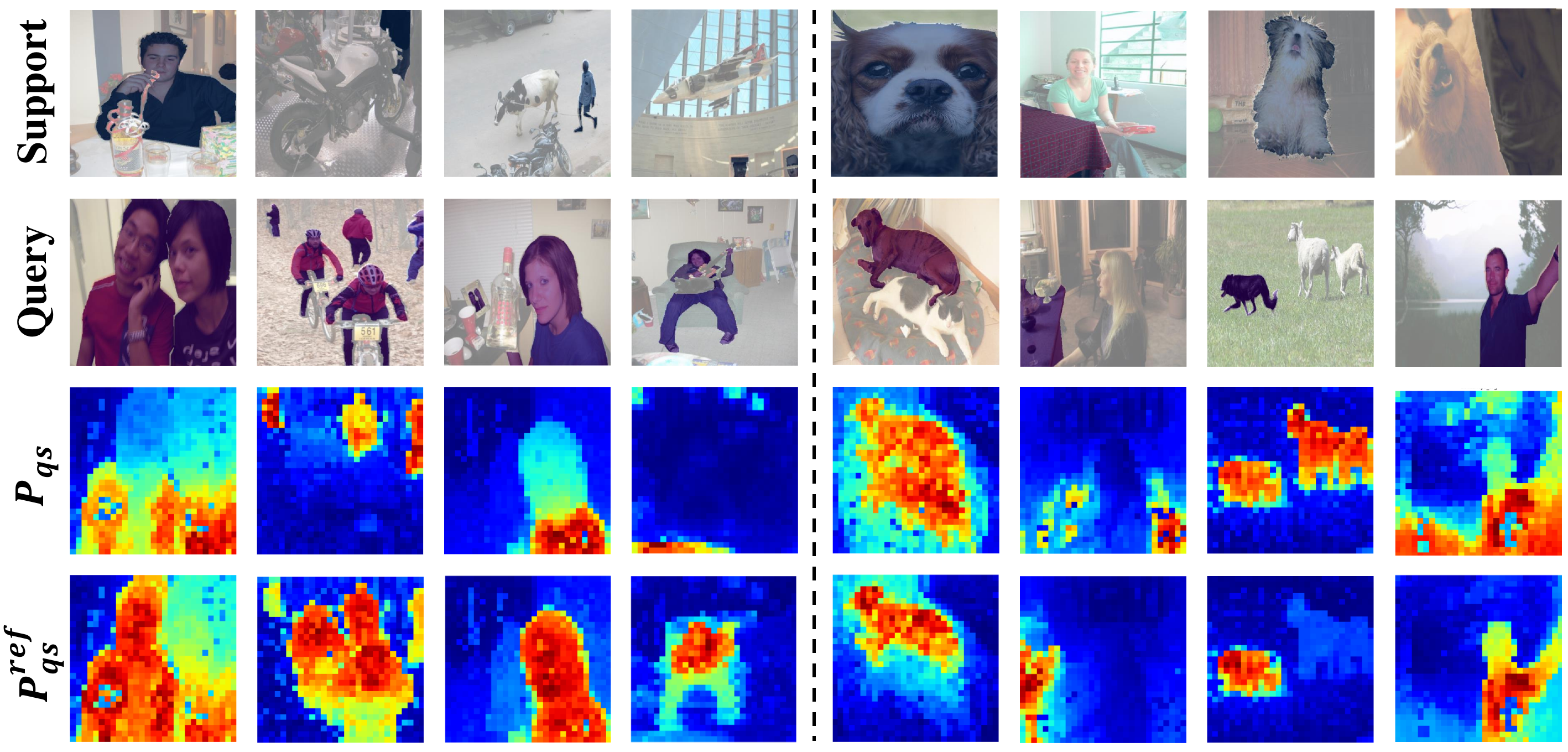}
    \caption{Qualitative comparison of visual priors. Each row (from top to bottom) represents: support images with ground-truth (GT) masks overlaid in blue, query images with GT masks in purple, visual priors $P_{qs}$ generated by previous methods, and the refined visual priors $P_{qs}^{ref}$ by our proposed MGTNet, which demonstrate more complete and accurate foreground activation.}
    \label{fig:motivation2}
  \end{figure}

Moreover, \textbf{most existing approaches treat the textual modality solely as a prior generator, without performing explicit cross-modal feature interaction or alignment}. As a result, they fail to fully exploit the semantic complementarity between visual and textual cues—especially in scenarios with large intra-class variations, where stronger semantic guidance is essential for producing reliable visual priors. As shown in Fig.~\ref{fig:motivation2}, the first four columns of $P_{qs}$ illustrate how large intra-class variations between the support and query images lead to incomplete activation of the foreground regions in the visual priors. 
Additionally, \textbf{the effectiveness of visual priors critically hinges on the semantic reliability of support foreground features.} In practice, certain foreground regions in support samples may exhibit low relevance to the target class but share high similarity with background or distractor classes—particularly near object boundaries where background pixels are mistakenly included. This may result in incorrect activation in background areas of the query image, ultimately degrading segmentation accuracy. 
As seen in the last four columns of $P_{qs}$ in Fig.~\ref{fig:motivation2}, background regions are falsely activated due to such interference. 

To overcome the aforementioned limitations, we propose the Multi-Text Guided Few-Shot Semantic Segmentation Network (MTGNet), as illustrated in Fig.~\ref{fig:motivation}(c).
MTGNet explicitly separates the refinement process into two dedicated branches:
(1) a textual prior refinement branch, which processes multi-text embeddings through the Multi-Textual Prior Refinement (MTPR) module. MTPR selectively propagates high-confidence signals and aggregates complementary cues across multiple descriptions, enhancing the completeness and reliability of foreground activation—particularly in complex scenes with high intra-class variations; and
(2) a visual prior refinement branch, which improves the quality of visual priors through two components. The Text Anchor Feature Fusion (TAFF) module reduces intra-class variations by treating multi-text embeddings as semantic anchors, guiding the fusion of semantically related regions between support and query features. This anchor-based consistency facilitates the transfer of fine-grained local prototypes, resulting in more discriminative query representations. Complementarily, the Foreground Confidence-Weighted Attention (FCWA) module enhances visual prior reliability by exploiting the internal self-similarity within support foreground features to suppress semantically inconsistent or irrelevant regions. This suppressive guidance is then propagated to the query image, improving segmentation robustness in cluttered or ambiguous scenes.

By incorporating diverse textual descriptions and enabling deep cross-modal feature interaction through its dual-branch design, MTGNet effectively addresses several key limitations of prior methods, including incomplete foreground activation, semantic misalignment between support and query, and the potential unreliability of local support features. These improvements ultimately lead to enhanced segmentation accuracy under limited supervision.

Our main contributions are summarized as follows:
\begin{enumerate}
    \item We propose the MTGNet, a novel dual-branch framework for few-shot segmentation that leverages multiple class-specific textual descriptions to construct richer and more complete visual and textual priors, overcoming the limitations of single-text-based approaches.
    \item We design three targeted modules to address different aspects of the segmentation challenge: an MTPR module that refines textual priors by aggregating and propagating complementary multi-text cues;
    a TAFF module that enhances support-query alignment through anchor-based visual-textual feature fusion; 
    and an FCWA module that improves visual prior reliability by suppressing semantically inconsistent support foreground regions. 
    \item Extensive experiments on standard FSS benchmarks demonstrate the effectiveness of MTGNet, achieving 76.8\% mIoU on $\text{PASCAL-}5^i$ and 57.4\% on $\text{COCO-}20^i$ in the 1-shot setting, with notable improvements on folds exhibiting high intra-class variations.
  \end{enumerate}
\section{Related Work}
\subsection{Few-shot Segmentation}
Few-shot semantic segmentation aims to segment novel categories with only a limited number of annotated samples. Traditional methods primarily adopt meta-learning paradigms, which can be broadly divided into prototype-based matching~\cite{zhang2020sg, fan2022self, cao2023break, yang2020prototype, li2021adaptive} and pixel-level matching approaches~\cite{hong2022cost, tian2020prior}. 

Prototype-based approaches aggregate support features into class-specific prototypes, which are then compared with query features via fixed or learnable similarity metrics. 
SG-One~\cite{zhang2020sg} introduces a widely adopted masked average pooling paradigm for prototype extraction, while SSP~\cite{fan2022self} refined prototypes by integrating high-confidence foreground regions from the query image. BBD~\cite{cao2023break} employed a transformation-and-fusion mechanism to strengthen the discriminative power of support prototypes.
However, single-prototype representations are often inadequate for capturing fine-grained semantics and intra-class variations. To address this, multi-prototype approaches have been proposed.
PMMs~\cite{yang2020prototype} used multiple prototypes to represent diverse semantic regions. ASGNet~\cite{li2021adaptive} applied superpixel-guided clustering to generate spatially adaptive prototypes, while PPNet~\cite{liu2020part} introduced part-aware prototypes to enhance spatial granularity and region coverage.

Pixel-level matching methods construct dense support-query correlations to guide segmentation directly at the pixel level. For example, VAT~\cite{hong2022cost} improved segmentation accuracy by aggregating multi-level correlation maps. PFENet~\cite{tian2020prior} generated a class-specific prior map by correlating high-level semantic features, which served as a soft mask for query segmentation.

\subsection{Language-Driven FSS}

The vision-language model CLIP~\cite{radford2021learning} aligns visual and textual data in a unified embedding space, enabling the use of rich semantic priors across modalities. This capability has been increasingly leveraged in FSS~\cite{luddecke2022image, leng2025multi, wang2024rethinking, chen2024visual}.

CLIPSeg~\cite{luddecke2022image} first applied CLIP to FSS by extracting text features through CLIP’s encoder and integrating them into the decoder via FiLM layers for pixel-level guidance.
For prior construction, Leng et al.\cite{leng2025multi} employed GradCAM to generate class activation maps, serving as weak localization cues.
PI-CLIP\cite{wang2024rethinking} proposed a dual-branch framework where GradCAM is also used in its vision-language branch. An additional refinement module further improves the quality of these priors, leading to enhanced performance.
PGMA-Net~\cite{chen2024visual} addressed base-class bias by aligning visual and textual features to build class-agnostic priors.

Most existing methods use a single text prompt (e.g., “a photo of \{class\}”) to represent the target category, overlooking the semantic diversity of complex classes with multiple subcomponents (e.g., faces, clothing, limbs). This often leads to incomplete foreground activation and reduced segmentation accuracy in complex scenarios. Unlike previous methods that simply fuse textual priors with visual features, our approach goes further by explicitly using multiple class-specific text prompts to enrich semantics and improve cross-modal alignment. Instead of treating text as a passive aid, we make it a core component for guiding feature refinement and fusion, enhancing both the completeness and consistency.
\section{Methodology}

\begin{figure*}[t!]
	\centering
	\includegraphics[width=0.9 \linewidth]{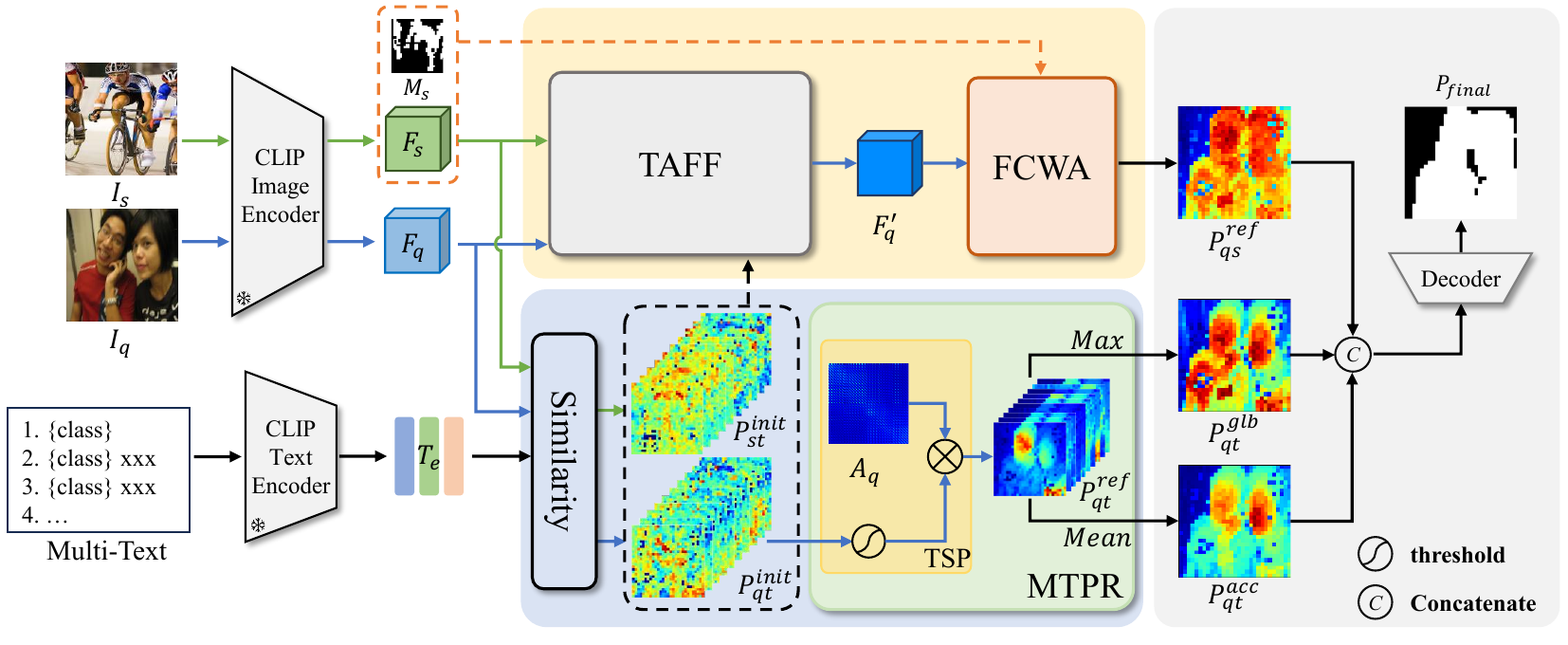}
        \caption{Overview of the proposed MTGNet pipeline. MTGNet adopts a dual-branch prior refinement strategy. 
        In the textual branch, the Multi-Textual Prior Refinement (MTPR) module refines $P_{qt}^{init}$ into $P_{qt}^{glb}$ and $P_{qt}^{acc}$ via threshold-based propagation and multi-text aggregation.
        In the visual branch, the Text Anchor Feature Fusion (TAFF) module extracts regional prototypes from $F_s$ guided by $P_{st}^{init}$ and aligns them with $F_q$ via $P_{qt}^{init}$.
        The Foreground Confidence-Weighted Attention (FCWA) module then generates a robust visual prior $P_{qs}^{ref}$. 
        Finally, the refined priors $P_{qt}^{glb}$, $P_{qt}^{acc}$, and $P_{qs}^{ref}$ are fused and decoded by an HDMNet-based~\cite{peng2023hierarchical} decoder to produce the final segmentation map $P_{final}$.
        }
	\label{fig:framework}
\end{figure*}

\subsection{Problem Setup}

Few-shot semantic segmentation generally relies on meta-learning and episodic sampling to acquire transferable segmentation knowledge from base classes during training, enabling the model to generalize to novel classes in the test stage with only a few support examples.
Specifically, during training, the model learns from a baseline dataset ${D}_b$ consisting of base classes $C_b$, and it is expected to perform an accurate inference on new classes ${C}_n$ from a novel dataset ${D}_n$ during testing. It is important to note that the base and novel classes are disjoint, i.e. ${C}_b \cap {C}_n = \emptyset$.
To facilitate this generalization, we follow the widely adopted episodic training paradigm~\cite{tian2020prior, wang2024rethinking, lin2023clip}, where both training and testing are conducted through episodic tasks. 
Each episode task consists of a support set ${S}$, a query set ${Q}$ and a textual description set $T$, where $T$ denotes the set of class-specific textual descriptions.

A typical episode follows the $N$-way $K$-shot configuration, meaning that $N$ distinct classes are sampled, each with $K$ annotated support images (typically $K=1$ or $5$). 
The limited visual labeled data forms the support set \( {S} = \{(I_{s}^i, M_{s}^i)\}_{i=1}^{N \times K} \), and the textual labeled data comes from the textual description set \( T = \{T_{e}^i\}_{i=1}^{N \times M} \), with \( M \) textual descriptions available per class, and the data to be segmented is the query set \( {Q} = \{(I_{q}^i, M_{q}^i)\}_{i=1}^{N} \). Here \( I_{s} \in \mathbb{R}^{H_{s} \times W_{s} \times 3} \), \( I_{q} \in \mathbb{R}^{H_{q} \times W_{q} \times 3} \), \( M_{s} \in \mathbb{R}^{H_{s} \times W_{s}} \), \( M_{q} \in \mathbb{R}^{H_{q} \times W_{q}} \), and \( T_{e} \in \mathbb{R}^{L} \) represent the support image, query image, the ground-truth masks of the support and query, and textual description, respectively. During the inference phase, the model makes predictions for the query image $ I_q $ by using the query-support pair $ \{I_q, S, T\} $, and evaluate the predicted segmentation output with the ground-truth mask of query image $M_q$.

\subsection{Multi-Text Guided Few-Shot Semantic Segmentation}

\subsubsection{Overview}

The pipeline of our proposed MTGNet, illustrated in Fig.~\ref{fig:framework}, consists of four main components: the Multi-Textual Prior Refinement (MTPR) module (Section~\ref{subsubsection:MTPR}), the Text Anchor Feature Fusion (TAFF) module (Section~\ref{subsubsection:TAFF}), the Foreground Confidence-Weighted Attention (FCWA) module (Section~\ref{subsubsection:FCWA}), and a decoder (Section~\ref{subsubsection:decoder}) that produces the final segmentation map.

Specifically, MTGNet adopts a dual-branch refinement strategy to refine both textual and visual priors. In the textual prior refinement branch, the MTPR module refines the initial text-query prior $P_{qt}^{init}$ via two mechanisms: a Threshold Similarity Propagation strategy that filters noisy guidance, and a Multi-Text Aggregation strategy that integrates complementary semantic cues from diverse textual descriptions. This results in two refined priors: a globally aggregated prior $P_{qt}^{glb}$ and an accurate, confidence-weighted prior $P_{qt}^{acc}$.
In the visual prior refinement branch, the TAFF module facilitates cross-sample semantic transfer by leveraging a visual-text-visual consistency strategy. It uses multi-text embeddings as semantic anchors to guide the propagation of discriminative prototypes from support image to semantically aligned regions in the query image, producing enhanced query features $F_q^{'}$. 
On the top of that, the FCWA module takes $F_q^{'}$, $F_s$ and $M_s$ as input, and leverages internal self-correlation to suppress noisy support regions as well as generates a reliable visual prior $P_{qs}^{ref}$ through cross-image similarity reasoning. Finally, the refined visual and textual priors are concatenated and passed through an HDMNet-based~\cite{peng2023hierarchical} decoder to produce the final segmentation map $P_{final}$. 

\subsubsection{Construction of Multi-Text Descriptions}

To improve foreground activation and semantic coverage, we construct a set of diverse textual descriptions for each class. Instead of relying on a single prompt, we generate multiple class-specific descriptions that vary in phrasing, attribute focus, and contextual cues. These descriptions are designed to attend to different semantically meaningful local regions of the target object (e.g., shape, parts, appearance), with the goal of covering the entire object rather than just its most prominent features.

\begin{figure}[t!]
	\centering
	\includegraphics[width=0.6\linewidth]{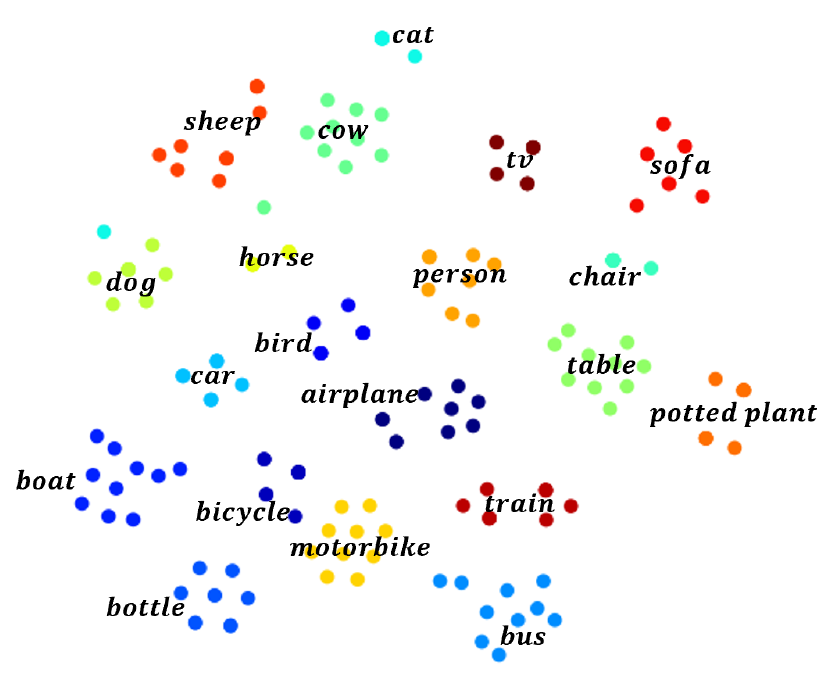}
	\caption{Visualization for the t-SNE~\cite{van2008visualizing} embeddings for the constructed  textual descriptions.}
	\label{fig:vistsne}
\end{figure}

Fig.~\ref{fig:vistsne} presents a t-SNE~\cite{van2008visualizing} visualization of the textual embeddings for the 20 categories in the PASCAL-5$^i$ dataset. As shown in Fig.~\ref{fig:vistsne}, the constructed multi-text embeddings form compact clusters within each class and are well-separated across classes, indicating strong intra-class consistency and inter-class discriminability. 

\subsubsection{CLIP-based Feature Extraction}
Given multiple textual descriptions for each category, we first use the CLIP text encoder to obtain the text embeddings $T_{e} \in \mathbb{R}^{n \times d}$, where $n$ is the number of descriptions and $d$ is the feature dimension. We extract visual features from both the query and support images by using the CLIP visual encoder, obtaining $F_q,F_s \in \mathbb{R}^{d \times hw}$, where $hw$ denotes the flattened spatial resolution of the feature map. For general reference, we denote them as $F_{visual}$, where $visual \in \{q, s\}$.

To establish the correspondence between visual and textual features, we compute cosine similarities between the text embeddings $T_{e}$ and visual features $F_{visual}$. This produces the initial textual-guided priors, denoted as $P_{qt}^{init}, P_{st}^{init} \in \mathbb{R}^{n \times hw}$, each indicating the relevance of every pixel to the corresponding textual prompts. For notation convenience, we refer to them generally as $P_{text}^{init}$, where $text \in \{qt, st\}$.
To ensure consistency across different descriptions, we apply min-max normalization to scale the similarity scores to the range \([0,1]\), forming a unified semantic prior:
\begin{equation}
	P_{text}^{init} = \operatorname{MinMaxNorm} \left( \frac{T_e F_{visual}}{\|T_e\| \|F_{visual}\|} \right).
\end{equation}

To capture internal semantic consistency within each image, we first extract self-attention weights $W_{visual} \in \mathbb{R}^{hw \times hw }$ from the ViT visual encoder.
These weights reflect relationships among spatial positions.
Then we apply the Sinkhorn normalization~\cite{sinkhorn1964relationship} to transform $W_{visual}$ into a symmetric self-correlation map $A_{visual} \in \mathbb{R}^{hw \times hw}$, i.e.,
\begin{equation}
	A_{visual} = \operatorname{Sinkhorn}\left(W_{visual}\right),  visual \in \{q, s\}.
\end{equation}
This normalization enforces spatial consistency and balance in the correlation, which is critical for downstream operations such as similarity propagation and foreground refinement.

\subsubsection{Multi-Textual Prior Refinement Module}
\label{subsubsection:MTPR}

To enhance the completeness and precision of semantic priors derived from multiple textual prompts, we propose the Multi-Textual Prior Refinement (MTPR) module, illustrated in Fig.~\ref{fig:framework}.  
The module performs refinement in two stages: Threshold Similarity Propagation (TSP), and Multi-text Aggregation.

\paragraph{Threshold Similarity Propagation}
Given the initial text-query prior $P_{qt}^{init} \in \mathbb{R}^{n \times hw}$, a thresholding operation is first applied to suppress unreliable activations and retain high-confidence foreground responses, resulting in the thresholded prior $P_{text}^{thr} \in \mathbb{R}^{n \times hw}$:
\begin{equation}
	P_{qt}^{thr}(j) = P_{qt}^{init}(j) \cdot \mathbf{1}_{\left\{ P_{qt}^{init}(j) \geq \tau \right\}},
\end{equation}
where $\mathbf{1}\{\cdot\}$ denotes the indicator function and $\tau=0.73$ is empirically determined.

To further enhance spatial consistency, the retained activations are propagated via the self-correlation map $A_q \in \mathbb{R}^{hw \times hw}$ computed from the query image:
\begin{equation}
	P_{qt}^{ref} = P_{qt}^{thr} A_{q}.
\end{equation}
This process encourages semantic continuity and expands activations to neighboring regions with strong feature affinities. The overall procedure can be compactly expressed as:
\begin{equation} 
	P_{qt}^{ref} = \operatorname{TSP}(P_{qt}^{init}, A_{q}). 
\end{equation}

\paragraph{Multi-text Aggregation}
To obtain comprehensive semantic guidance, we aggregate the refined priors $P_{qt}^{ref}$ from all textual descriptions. Specifically, we compute both a global prior via max pooling and a confidence-weighted prior via average pooling, applied across the textual dimension $n$ (i.e., for each spatial location, values across all $n$ textual prompts are aggregated):
\begin{equation}
	P_{qt}^{glb} = \operatorname{Reshape} \left( \operatorname{Max} \left(P_{qt}^{ref} \right) \right),
\end{equation}
\begin{equation}
	P_{qt}^{acc} = \operatorname{Reshape} \left( \operatorname{Mean} \left(P_{qt}^{ref} \right)\right),
\end{equation}
where $\operatorname{Reshape}(\cdot)$ converts each $hw$-length vector into a 2D spatial map of size $h \times w$. $P_{qt}^{glb}$ emphasizes the most salient activations across all prompts, promoting broad semantic coverage, while $P_{qt}^{acc}$ offers a more balanced and noise-suppressed prior. Together, these outputs provide robust and complementary guidance for query segmentation.

\subsubsection{Text-Anchor Feature Fusion}
\label{subsubsection:TAFF}
\begin{figure}
	\centering
	\includegraphics[width=0.9\linewidth]{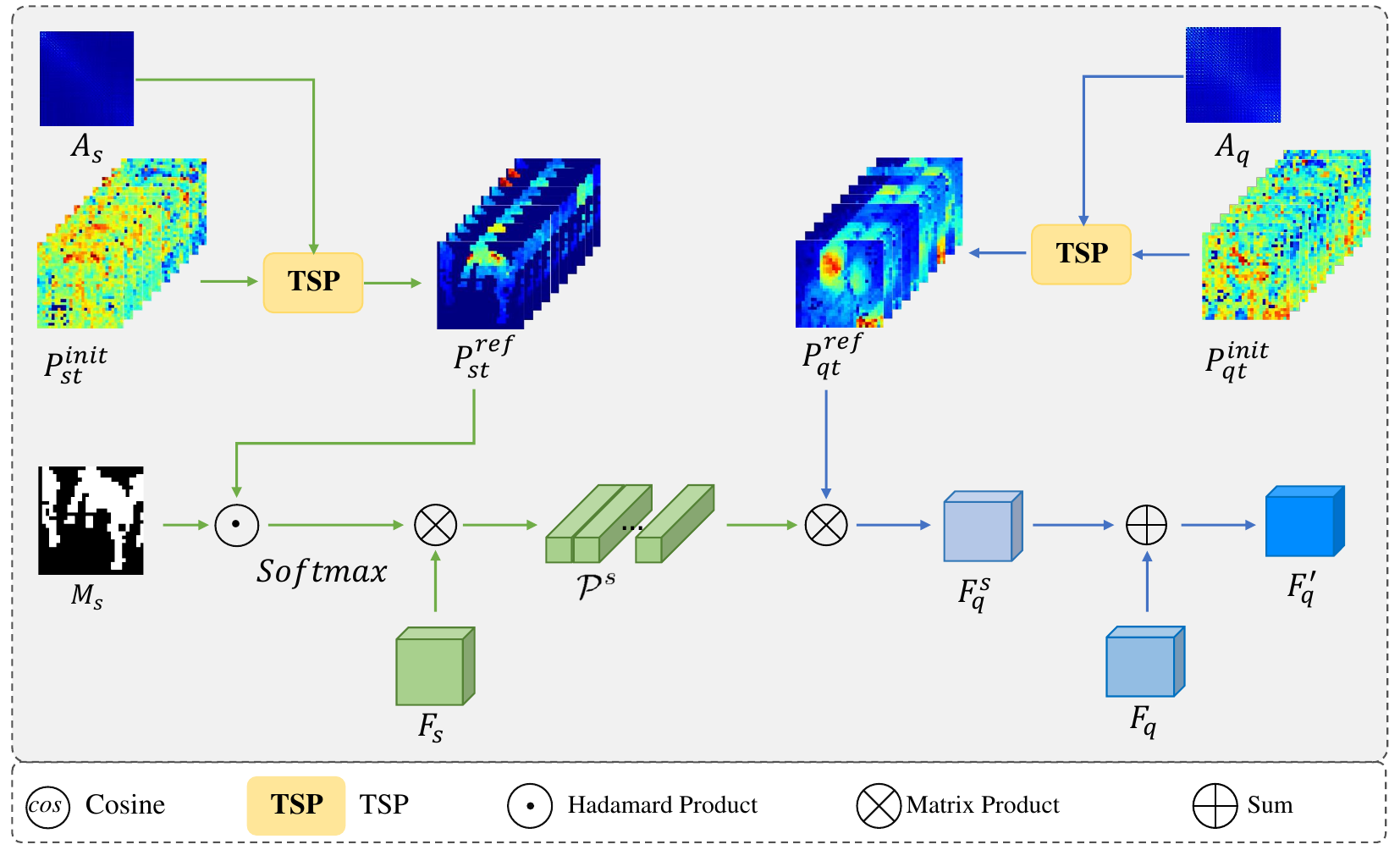}
	\caption{The structure of our proposed Text-Anchor Feature Fusion (TAFF) module.}
	\label{fig:TAFF}
        \vspace{-3pt}
\end{figure}

To address intra-class variations and enhance alignment between support and query features, we propose the Text-Anchor Feature Fusion (TAFF) module, as shown in Fig.~\ref{fig:TAFF}. TAFF implements a visual-text-visual consistency strategy to facilitate semantic transfer from support to query by leveraging multi-text anchors. It uses refined multi-text priors from both support and query to guide the alignment and fusion of semantically relevant features, ultimately enriching the query representation while maintaining structural coherence.

The initial priors $P_{st}^{init}$ and $P_{qt}^{init}$ are first refined via the TSP operation, resulting in $P_{st}^{ref}$ and $P_{qt}^{ref}$, respectively. 
Given the support feature $F_{s}$, the corresponding support mask $M_{s} \in \mathbb{R}^{1 \times {hw}}$, and the refined text-support priors $P_{st}^{ref} \in \mathbb{R}^{n \times hw}$, 
a set of regional prototypes $\mathcal{P}^{s} \in \mathbb{R}^{n \times d}$ is computed by aggregating masked support features weighted by the attention scores from $P_{st}^{ref}$:
\begin{equation}
	\mathcal{P}^{s} = \operatorname{softmax} \left( P_{st}^{ref} \odot M_{s} \right) F_{s}^{T},
\end{equation}
where $\odot$ denotes the Hadamard product, and $\operatorname{softmax}(\cdot)$ is applied along the spatial dimension $hw$ to normalize attention across spatial locations. In this way, each row $\mathcal{P}^s(i, :)$ serves as a semantic prototype correspondign to the $i$-th textual anchor, capturing diverse part-level characteristics of the target object.

To facilitate semantic transfer across images, the refined query-text prior $P_{qt}^{ref} \in \mathbb{R}^{n \times hw}$ is used to inject the support prototypes into semantically relevant regions of the query image, generating an enhanced query representation:
\begin{equation}
	F_{q}^{s} =  \left( P_{qt}^{ref}  \right)^{T} \mathcal{P}^{s}.
\end{equation}

Subsequently, this enhanced query representation is fused with the original query feature $F_q$ to produce the refined query feature $F_q' \in \mathbb{R}^{d \times hw}$:
\begin{equation}
	F_{q}^{'} =  \left( F_{q}^{s}  \right)^{T} + F_{q}.
\end{equation}
This fusion integrates multi-text semantic guidance with the original query features, enabling the model to capture richer discriminative cues and exhibit greater robustness against intra-class variations.

\subsubsection{Foreground Confidence-Weighted Attention}
\label{subsubsection:FCWA}

\begin{figure}
	\centering
	\includegraphics[width=0.9 \linewidth]{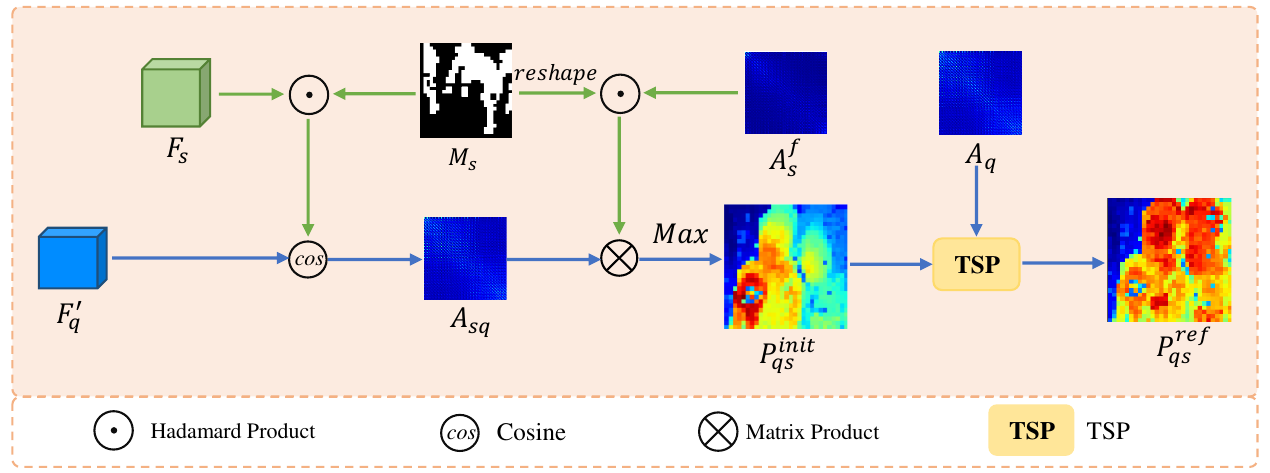}
	\caption{The structure of our proposed Foreground Confidence-Weighted Attention (FCWA) module.}
	\label{fig:FCWA}
\end{figure}

The Foreground Consistency Weighted Attention (FCWA) module enhances the reliability of visual priors by suppressing interference within support foreground and reinforcing query activation through semantic consistency. As illustrated in Fig.~\ref{fig:FCWA}, FCWA exploits the internal self-correlation of the support foreground to highlight those reliable regions and downweight noisy ones, guiding accurate target localization in the query image.

Given the support feature $F_{s}$ and its corresponding mask $M_{s}$, we first extract the support foreground feature $F_{s}^{'} \in \mathbb{R}^{d \times hw}$ via element-wise multiplication:
\begin{equation}
	F_{s}^{'} =  F_{s} \odot M_{s}.
\end{equation}

We then compute the cosine similarity between $F_{s}^{'}$ and the refined query features $F_{q}^{'}$ to obtain the cross-correlation map $A_{sq} \in \mathbb{R}^{hw \times hw}$:
\begin{equation}
	A_{sq} =  \frac{(F_{s}^{'})^{T} F_{q}^{'}}{\|F_{s}^{'}\| \|F_{q}^{'}\|}.
\end{equation}

To assess consistency within the support foreground, we refine the support self-correlation map $A_{s}$ of the support image by using the mask $M_{s}$ to form $A_{s}^{f}\in \mathbb{R}^{hw \times hw}$: 
\begin{equation}
	A_{s}^{f} =   A_{s}  \odot (M_{s})^{T}.
\end{equation}

The self-correlation map $A_s^f$ is used to reweight $A_{sq}$, yielding a consistency-aware cross-image correlation map, i.e.,
\begin{equation} 
	A_{sq}^{'} = A_{s}^{f} A_{sq}. 
\end{equation}

A max pooling operation is then applied to extract the most salient activations from $A_{sq}'$, producing the initial visual prior $P_{qs}^{init} \in \mathbb{R}^{1 \times hw}$:
\begin{equation} 
	P_{qs}^{init} = \operatorname{Max}(A_{sq}^{'}).
\end{equation}

Although max pooling effectively highlights dominant regions, it may also inadvertently emphasize spurious background activations caused by overly confident but incorrect matches. To address this, the TSP operation is applied to suppress noisy activations and enhance spatial coherence:
\begin{equation} 
	P_{qs}^{ref} = \operatorname{Reshape}( \operatorname{TSP}(P_{qs}^{init}, A_{q})). 
\end{equation}
The resulting $P_{qs}^{ref}$ serves as a more accurate and robust visual prior to guide query segmentation.

\subsubsection{Decoding and Final Prediction}
\label{subsubsection:decoder}
To generate the final segmentation output, the previously obtained multimodal-guided priors are used to guide a more detailed and accurate prediction. We leverage the HDMNet decoder~\cite{peng2023hierarchical}, which features a hierarchical attention architecture well-suited for modeling complex object structures. In addition, our design draws inspiration from the text-guided few-shot segmentation strategy proposed in PI-CLIP~\cite{wang2024rethinking}, which effectively integrates textual semantics into the decoding process.  
Specifically, the textual-guided priors $P_{qt}^{glb}$ and $P_{qt}^{acc}$,  along with the visual-guided prior $P_{qs}^{ref}$, are concatenated to form a unified multimodal representation. This fused prior encapsulates complementary semantic cues from different modalities. The decoder then takes this enriched representation as input and produces the final segmentation map:
\begin{equation} 
	P_{final} = \operatorname{Decoder}( \operatorname{Concat}(P_{qt}^{glb},P_{qt}^{acc},P_{qs}^{ref}) ). 
\end{equation}

To supervise the training process, we adopt the Binary Cross-Entropy (BCE) loss, commonly used in binary semantic segmentation tasks. Let $Q \in \mathbb{R}^{H \times W \times 2}$ denotes the model’s predicted foreground-background probability map, and $M \in \mathbb{R}^{H \times W}$ be the corresponding binary ground truth mask. The BCE loss quantifies the pixel-wise discrepancy between predictions and ground truths, and guides the model’s parameter updates by penalizing incorrect predictions across all pixels. It is formally defined as: 
\begin{equation}
\mathcal{L}_{CE} = - \sum_{i=1}^{HW} \bigl[ M(i) \log Q_1(i) + (1 - M(i)) \log Q_0(i) \bigr],
\end{equation}
where $Q_1(i)$ and $Q_0(i)$ represent the predicted probabilities for the foreground and background classes at pixel $i$, respectively, and $M(i) \in \{0, 1\}$ is the binary ground truth label.
\section{{Experiments}}

\subsection{Experimental Setup}
\subsubsection{Datasets}
We evaluate our method on two standard FSS benchmarks: $\text{PASCAL-}5^i$~\cite{shaban2017one} and $\text{COCO-}20^i$~\cite{nguyen2019feature}.
The $\text{PASCAL-}5^i$ dataset is constructed from the PASCAL VOC 2012 dataset~\cite{everingham2010pascal}, with additional semantic annotations sourced from the SBD dataset~\cite{hariharan2011semantic}. It includes 20 object classes split into four folds of five categories each. In each evaluation, one fold is used for testing, while the other three serve for training. This cross-validation ensures balanced and comprehensive assessment by rotating test folds.
The $\text{COCO-}20^i$ dataset, based on MS COCO~\cite{lin2014microsoft}, offers a more challenging benchmark with 80 categories and complex scenes. It is divided into four folds of 20 classes each. Following the standard protocol~\cite{nguyen2019feature}, three folds are for training, and one for testing. This strict class-level split ensures evaluation on unseen categories, testing the model’s generalization ability.

\subsubsection{Evaluation Metrics}
For quantitative performance evaluation, we adopt two standard metrics: mean Intersection over Union (mIoU) and Foreground-Background Intersection over Union (FB-IoU), consistent with prior FSS studies.

IoU is defined as the ratio of the intersection to the union of the predicted and ground-truth segmentation masks for a particular class: 
$
\text{IoU} = \frac{\text{TP}}{\text{TP} + \text{FP} + \text{FN}},
$
where TP, FP, and FN represent the numbers of true positive, false positive, and false negative pixels, respectively.
The mIoU is computed by averaging the IoU scores over all foreground classes within a fold:
$\text{mIoU} = \frac{1}{N} \sum_{c=1}^{N} \text{IoU}_c,$
where $N$ is the number of foreground classes and $\text{IoU}_c$ denotes the IoU score for class $c$. This metric offers an overall measure of segmentation accuracy across categories.
In addition, we employ FB-IoU to provide a balanced view of segmentation quality over both foreground and background regions. It is computed as the average of foreground IoU ($\text{IoU}_{\text{fg}}$) and background IoU ($\text{IoU}_{\text{bg}}$), i.e.,
$\text{FB-IoU} = \frac{1}{2}(\text{IoU}_{\text{fg}} + \text{IoU}_{\text{bg}}).$

\subsection{Implementation Details}
All experiments are performed on the $\text{PASCAL-}5^i$ and $\text{COCO-}20^i$ benchmarks, with input images uniformly resized to $473 \times 473$ pixels. 
We utilize the CLIP ViT-B/16 model~\cite{radford2021learning} as a frozen pre-trained backbone for feature extraction.
For fair comparison, we adopt the same configurations for data augmentation, learning rate and optimizer as used in the baseline model HDMNet~\cite{min2021hypercorrelation}. 
All training is conducted on a single NVIDIA GeForce RTX 3090 GPU using an episodic training paradigm. Specifically, the model is trained for 150 epochs on $\text{PASCAL-}5^i$ and 50 epochs on $\text{COCO-}20^i$, with a batch size of 8 for the 1-shot and 4 for the 5-shot settings.

During inference, no data augmentation is applied, and the batch size is fixed to 1. To ensure reproducibility and fair evaluation, we follow the testing protocol of prior works~\cite{wang2024rethinking, peng2023hierarchical}, performing 10 independent runs per fold, each with a distinct random seed. In each test round, 1,000 query-support pairs are sampled for $\text{PASCAL-}5^i$ and 5,000 pairs for $\text{COCO-}20^i$.

\subsection{Comparison with State-of-the-Art Methods}

Table~\ref{tab:pascal_sota} summarizes the performance on the $\text{PASCAL-}5^i$ benchmark. Our MTGNet consistently surpasses conventional FSS methods that do not incorporate textual guidance. Compared to the baseline HDMNet~\cite{min2021hypercorrelation}, our model achieves substantial gains of 7.4\% and 6.5\% in mIoU for the 1-shot and 5-shot settings, respectively.
When benchmarked against CLIP-based methods that utilize single textual prompts, MTGNet demonstrates notable improvements across most folds by using the same CLIP-ViT-B/16 backbone. Specifically, compared to PGMA-Net~\cite{chen2024visual}, MTGNet achieves 2.7\% and 3.8\% higher mIoU in the 1-shot and 5-shot settings. Against PI-CLIP~\cite{wang2024rethinking}, our method further improves mIoU by 1.2\% in the 5-shot case. 
Our model is particularly effective on challenging folds. For example, on Fold 2, which contains complex categories such as person and table, MTGNet outperforms PGMA-Net by 9.2\% and 10.4\%, and surpasses PI-CLIP by 1.3\% and 2.4\% in the 1-shot and 5-shot scenarios, respectively.

\begin{table*}
	\centering
	\caption{Comparison of the proposed MTGNet with the current SOTA on $\text{PASCAL-}5^i$ ~\cite{shaban2017one}. Results in \textbf{bold} denote the best performance, while the \underline{underlined} ones indicate the second best.}
	\label{tab:pascal_sota}
	\scalebox{0.85}{
		\begin{tabular}{cclc|ccccc|ccccc}
			\toprule
			\multirow{2}{*}{\shortstack{\textbf{Pretrain}}} & \multirow{2}{*}{\shortstack{\textbf{Backbone}}} & \multirow{2}{*}{\textbf{Method}} & \multirow{2}{*}{\textbf{Publication}} & \multicolumn{5}{c}{\textbf{1-shot}} & \multicolumn{5}{c}{\textbf{5-shot}}  \\ 
			
			& & & & $\textbf{Fold0}$ & $\textbf{Fold1}$ & $\textbf{Fold2}$ & $\textbf{Fold3}$ & \textbf{Mean} & $\textbf{Fold0}$ & $\textbf{Fold1}$ & $\textbf{Fold2}$ & $\textbf{Fold3}$ & \textbf{Mean}  \\ 
			\midrule
	
			\multirow{8}{*}{IN1K} & \multirow{8}{*}{RN50} 
			
			& PFENet~\cite{tian2020prior} & TPAMI'20  & 61.7 & 69.5 & 55.4 & 56.3 & 60.8  & 63.1 & 70.7 & 55.8 & 57.9 & 61.9   \\ 
			
			& & HSNet~\cite{min2021hypercorrelation} & ICCV'21   & 64.3 & 70.7 & 60.3 & 60.5 & 64.0  & 70.3 & 73.2 & 67.4 & 67.1 & 69.5   \\
			
			& & SSP~\cite{fan2022self}  & ECCV'22   & 60.5 & 67.8& 66.4 & 51.0& 61.4 & 67.5 & 72.3 & 75.2 & 62.1 & 69.3  \\
			
			& & HPA\cite{cheng2023hpa} & TPAMI'23  & 65.9 & 72.0 & 64.7 & 56.8 & 64.8  & 70.5 & 73.3 & 68.4 & 63.4 & 68.9   \\  
			
			& & FECANet~\cite{liu2023fecanet} & TMM'23  & 69.2 & 72.3 & 62.4 & 65.7 & 67.4  & 72.9 & 74.0 & 65.2 & 67.8 & 70.0  \\
			
			& & HDMNet~\cite{peng2023hierarchical} & CVPR'23 & 71.0 & 75.4 & 68.9 & 62.1 & 69.4 & 71.3 & 76.2 & 71.3 & 68.5 & 71.9\\ 
            
                & & QPENet~\cite{cong2024query} & TMM'24 & 65.2 & 71.9 & 64.1 & 59.5 & 65.2 & 68.4 & 74.0 & 67.4 & 65.2 & 68.8 \\ 
            
			& & DGFPNet~\cite{wen2024dual} & TMM'24 & 70.5 & 74.5 & 70.0 & 61.9 & 69.2 & 74.0 & 76.3 & 72.3 & 67.5 & 72.5 \\ 
			
			\midrule 
                \multirow{5}{*}{CLIP} 

			&\multirow{5}{*}{CLIP-ViT-B/16}&  CLIPSeg~\cite{luddecke2022image} & CVPR'22 & - & - & - & - & 59.5  & - & - & - & - & -  \\
			
			& & PGMA-Net~\cite{chen2024visual}  &TMM'24 & 74.0 & 81.9 & 66.8 & \textbf{73.7} & \underline{74.1}  & 74.5 & 82.2 & 67.2 & \textbf{74.4} & 74.6  \\
			
			& & PI-CLIP~\cite{wang2024rethinking}  &CVPR'24 &   \underline{76.4} & \textbf{83.5} & \underline{74.7} & \underline{72.8} &  \textbf{76.8} & \underline{76.7} & \textbf{83.8} & 75.2 & 73.2 & \underline{77.2} \\ 

                & & ~\cite{leng2025multi} & TCSVT'25  & 74.6 & 79.6 & 74.5 & 69.0 & 74.4 & 75.3 & 80.4 & \underline{75.4} & 71.0 & 75.5 \\
       
			\cline{3-14} \\[-2.0ex] 

			&& MTGNet (ours)  &- & \textbf{77.7} &  \underline{82.2} &  \textbf{76.0} &  71.4 &  \textbf{76.8} &  \textbf{79.1} &  \underline{82.7} &  \textbf{77.6} &  \underline{74.0} &  \textbf{78.4}  \\        
			
			\bottomrule
			
		\end{tabular}
	}
	\hfill
\end{table*}

\begin{table*}[t]
	\centering
	\caption{Comparison of the proposed MTGNet with the current SOTA on $\text{COCO-}20^i$ dataset~\cite{nguyen2019feature}. Results in \textbf{bold} denote the best performance, while the \underline{underlined} ones indicate the second best.}
	\label{tab:coco_sota}
	\scalebox{0.85}{
		\begin{tabular}{cclc|ccccc|ccccc}
			\toprule
			\multirow{2}{*}{\shortstack{\textbf{Pretrain}}} & \multirow{2}{*}{\shortstack{\textbf{Backbone}}} & \multirow{2}{*}{\textbf{Method}} & \multirow{2}{*}{\textbf{Publication}} & \multicolumn{5}{c}{\textbf{1-shot}} & \multicolumn{5}{c}{\textbf{5-shot}}  \\ 
			& & & & $\textbf{Fold0}$ & $\textbf{Fold1}$ & $\textbf{Fold2}$ & $\textbf{Fold3}$ & \textbf{Mean} & $\textbf{Fold0}$ & $\textbf{Fold1}$ & $\textbf{Fold2}$ & $\textbf{Fold3}$ & \textbf{Mean}  \\ 
			\midrule
			
			\multirow{8}{*}{IN1K} & \multirow{8}{*}{RN50}
			& PFENet~\cite{tian2020prior} & TPAMI'20   & 36.5 & 38.6 & {34.5} & {33.8} & {35.8}  & 36.5 & 43.3 & 37.8 & 38.4 & 39.0   \\ 
			
                & & HSNet~\cite{min2021hypercorrelation} & ICCV'21    & 36.3 & 43.1 & 38.7 & 38.7 & 39.2  & 43.3 & 51.3 & 48.2 & 45.0 & 46.9  \\
    		
                & & SSP~\cite{fan2022self}  & ECCV'22   & 35.5 & 39.6& 37.9 & 36.7& 37.4 & 40.6 & 47.0 & 45.1 & 43.9 & 44.1   \\
    		
                & & HPA~\cite{cheng2023hpa} & TPAMI'23  & 40.3 & 46.6 & 44.1 & 42.7 & 43.4  & 45.5 & 55.4 & 48.9 & 50.2 & 50.0   \\  
    		
                & & FECANet~\cite{liu2023fecanet} & TMM'23  & 38.5 & 44.6 & 42.6 & 40.7 & 41.6  & 44.6 & 51.5 & 48.4 & 45.8 & 47.6    \\  
    		
                & & HDMNet~\cite{peng2023hierarchical} & CVPR'23 &43.8 &55.3 &51.6 &49.4 &50.0 &50.6 &61.6 &55.7 &56.0 &56.0\\ 
    		
                & & QPENet~\cite{cong2024query} & TMM'24 & 41.5 & 47.3 & 40.9 & 39.4 & 42.3 & 47.3 & 52.4 & 44.3 & 44.9 & 47.2 \\ 
    		
                & & DGFPNet~\cite{wen2024dual} & TMM'24 & 43.1 & 56.1 & 48.0 & 48.0 & 48.8 & 48.4 & 61.8 & 54.2 & 53.0 & 54.4 \\ 

			\midrule 
			
			\multirow{5}{*}{CLIP} 

                & CLIP-RN50& PGMA-Net~\cite{chen2024visual}  &TMM'24 &  \textbf{49.9} & 56.7 & \underline{55.8} & 54.7 & 54.3 & 49.5 & 61.7 & \underline{59.1}
            & 57.9 & 57.1 \\

                \cline{2-14} \\[-2.0ex]
                
                &\multirow{4}{*}{CLIP-ViT-B/16}&  CLIPSeg~\cite{luddecke2022image} & CVPR'22 & - & - & - & - & 33.3 & - & - & - & - & - \\ 

			& & PI-CLIP~\cite{wang2024rethinking}  &CVPR'24 &  49.3 & \textbf{65.7} & \underline{55.8} & \underline{56.3} & \underline{56.8}  & \textbf{56.4} & \textbf{66.2} & 55.9 & \underline{58.0} & \underline{59.1}  \\

                & & ~\cite{leng2025multi} & TCSVT'25  & \underline{49.8} & 61.0 & 53.2 & 54.5 & 54.6 & \underline{56.3} & 64.3 & 55.7 & 57.9 & 58.5 \\
            
			\cline{3-14} \\[-2.0ex]
			
			& & MTGNet (ours)  &- & 49.7 & \underline{63.1} & \textbf{58.3} & \textbf{58.3} & \textbf{57.4}  & \textbf{56.4} & \underline{65.6} & \textbf{61.7} & \textbf{60.3} & \textbf{61.0}  \\
			
			\bottomrule
		\end{tabular}
	}
	\hfill
\end{table*}

Table~\ref{tab:coco_sota} reports the results on the more challenging $\text{COCO-}20^i$ dataset. MTGNet delivers strong performance, achieving 7.4\% and 5.0\% improvements in mIoU over the HDMNet baseline for 1-shot and 5-shot tasks. Furthermore, it outperforms PI-CLIP by 0.6\% and 1.9\% in the same settings while using the identical CLIP-ViT-B/16 backbone.

\begin{figure*}[ht]
	\centering
	\includegraphics[width=0.8 \linewidth]{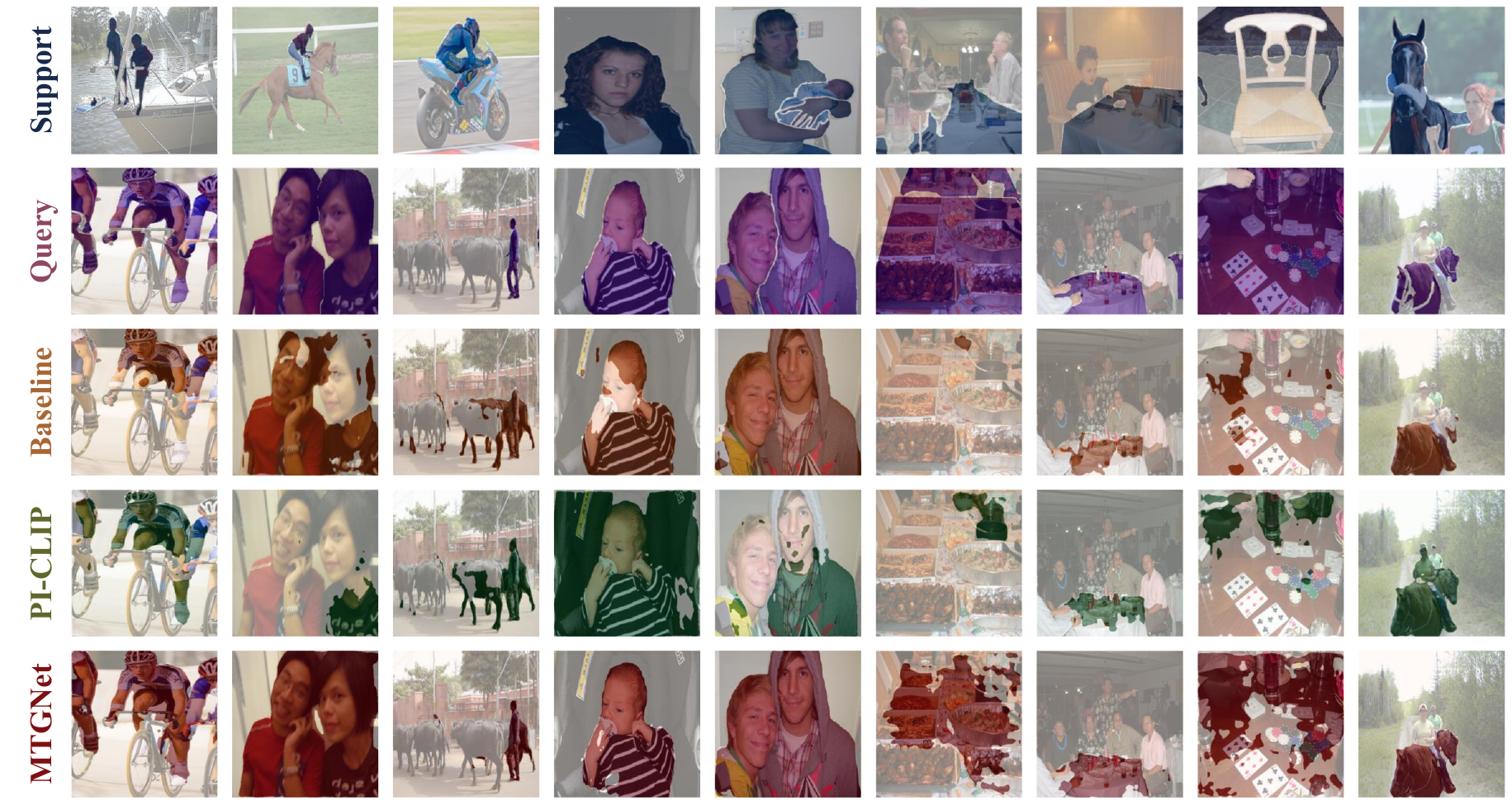}
	\caption{Qualitative comparison of segmentation results between the baseline (HDMNet), PI-CLIP, and our proposed MTGNet. Each column represents a different test sample, and each row (from top to bottom) shows: (1) the support image with ground-truth (GT) mask overlaid in blue, (2) the query image with GT mask in purple, (3) the baseline prediction in orange, (4) the PI-CLIP prediction in green, and (5) the MTGNet prediction in red. Our method consistently yields more complete and accurate segmentation results compared to the baseline and PI-CLIP.}
	\label{fig:vis}
    \vspace{-5pt}
\end{figure*}

To further validate the effectiveness of our proposed model, Fig.~\ref{fig:vis} showcases qualitative comparisons among the baseline (HDMNet), PI-CLIP, and our MTGNet across representative examples. As illustrated, our method consistently produces more accurate and complete segmentation masks compared to the baseline and PI-CLIP.

\begin{figure}[H]
	\centering
	\includegraphics[width=0.95 \linewidth]{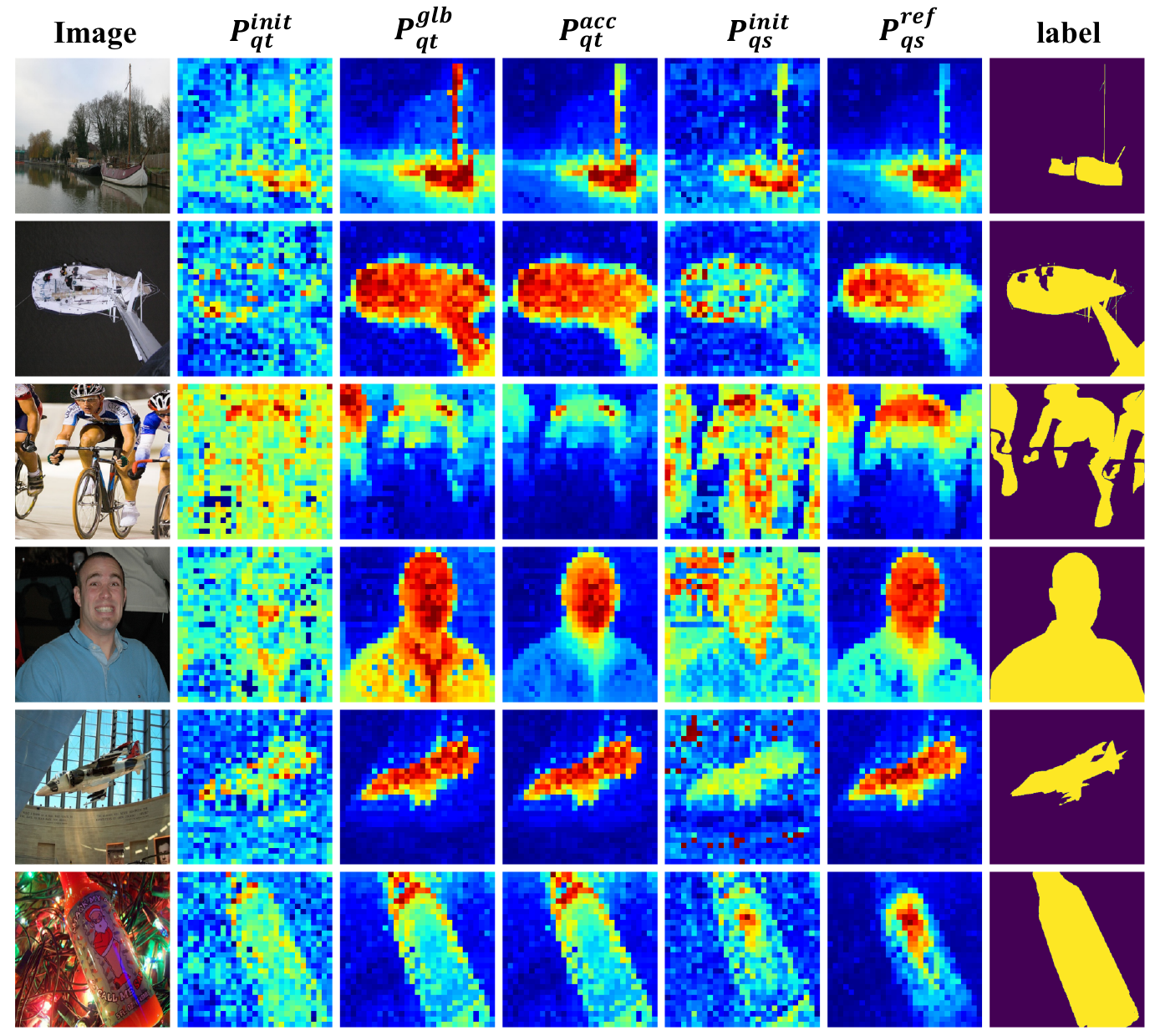}
	\caption{Visualization of various types of prior information used in MTGNet. Each row corresponds to a query sample, and each column (from left to right) shows: (1) the input query image, (2) the initial text-guided prior $P_{qt}^{{init}}$, (3) the comprehensive prior $P_{qt}^{{glb}}$ obtained by aggregating multiple textual prompts, (4) the high-confidence prior $P_{qt}^{{acc}}$ focusing on common semantic cues, (5) the initial visual-guided prior $P_{qs}^{{init}}$, (6) the refined visual prior $P_{qs}^{{ref}}$ generated by the FCWA module, and (7) the ground-truth label. These visualizations highlight how the proposed modules collaboratively enhance the relevance and precision of prior information, progressively refining spatial guidance for accurate segmentation.}
	\label{fig:visim}
\end{figure}

In addition, Fig.~\ref{fig:visim} provides visualizations of intermediate priors, offering insights into how each module contributes to refining segmentation guidance. The prior $P_{qt}^{glb}$, which aggregates semantic cues from diverse textual prompts, exhibits broad activation across target regions but may introduce some irrelevant responses. Conversely, $P_{qt}^{acc}$, computed by emphasizing the commonality among multiple prompts, focuses on high-confidence regions with less noise. Furthermore, the transformation of the initial visual-guided prior $P_{qs}^{init}$ into the refined version $P_{qs}^{ref}$ demonstrates the model’s ability to suppress background noise while enhancing the alignment with true object regions.

\subsection{Ablation Studies}
To assess the contribution of each component in our proposed MTGNet, we conducted detailed ablation experiments using HDMNet\cite{peng2023hierarchical} as the baseline. All ablations were carried out on Fold 0 of the $\text{PASCAL-}5^i$ dataset\cite{shaban2017one}.

\subsubsection{Impact of individual modules and their combinations} 
Each module in MTGNet is designed to generate distinct gui-
\noindent dance information. To evaluate the effectiveness of these components, we progressively integrated them into the baseline and reported the corresponding performance gains, as summarized in Table~\ref{tab:module_comparison}. 
The first row shows the baseline performance without any textual priors or additional modules. 
Introducing textual guidance and processing it through the MTPR module leads to a significant performance gain of 3.8\%, demonstrating the value of enriched semantic priors from diverse textual prompts.
Adding the TAFF module further improves performance by 1.1\%, indicating that aligning textual anchors with visual features enhances semantic consistency.
Incorporating the FCWA module yields an additional 1.84\% improvement by suppressing irrelevant support features and refining the visual prior.
These results confirm that each module contributes positively to the overall segmentation accuracy and that their combination leads to a substantial cumulative improvement over the baseline.

\begin{table}[t]
	\centering
	\caption{Comparison of MTPR, TAFF, FCWA, and their combinations on mIoU and FB-IoU.}
	\label{tab:module_comparison}
        \scalebox{0.85}{
	\begin{tabular}{cccccc}
		\toprule
		\textbf{baseline} & \textbf{MTPR} & \textbf{TAFF} & \textbf{FCWA} & \textbf{mIoU (\%)} & \textbf{FB-IoU (\%)} \\
		\midrule
		\checkmark &  &  &  & 71.00 & 85.86 \\
		\checkmark & \checkmark &  &  & 74.77 & 86.25 \\
		\checkmark & \checkmark & \checkmark & &  75.87 & 87.04 \\
		\checkmark & \checkmark & \checkmark & \checkmark &  \textbf{77.71} & \textbf{87.93} \\
		\bottomrule
	\end{tabular}
        }
        \vspace{-5pt}
\end{table}

\begin{table}[t]
	\centering
	\caption{Comparison of no-text, single-text, and multi-text on mIoU and FB-IoU.}
	\label{tab:text_comparison}
        \scalebox{0.85}{
	\begin{tabular}{ccccc}
		\toprule
		\textbf{No-Text} & \textbf{Single-Text} & \textbf{Multi-Text} & \textbf{mIoU (\%)} & \textbf{FB-IoU (\%)} \\
		\midrule
		\checkmark &  &  & 72.71 & 84.58 \\
		& \checkmark &  & 77.25 & 87.68 \\
		& & \checkmark & \textbf{77.71} & \textbf{87.93} \\
		\bottomrule
	\end{tabular}
        }
        \vspace{-5pt}
\end{table}

\begin{table}[t]
	\centering
	\caption{Comparison of single-text and multi-text across different categories.}
	\label{tab:text_comparison2}
        \scalebox{0.85}{
	\begin{tabular}{ccc}
		\toprule
		\textbf{Class} & \textbf{Single-Text mIoU (\%)} & \textbf{Multi-Text mIoU (\%)}  \\
		\midrule
		  table & 45.59 & 47.05 \\
		  person & 62.83 & 72.56 \\
            potted plant & 55.18 & 57.30 \\
            sofa  & 73.03 & 75.12 \\
		train & 76.98 & 83.51 \\
		\bottomrule
	\end{tabular}
        }
        \vspace{-5pt}
\end{table}

\subsubsection{Ablation study on textual information} 

To evaluate our textual guidance strategy, we compare three variants with identical architectures: (1) no text input, (2) single-text prompts (e.g., “{class}”), and (3) multi-text prompts. As shown in Table~\ref{tab:text_comparison}, removing textual input still improves performance by 1.7\% over the baseline, suggesting our structural enhancements are effective. Adding single-text prompts yields a further 4.54\% gain, highlighting the value of basic semantic priors. Multi-text prompts provide an additional 0.46\% boost, though the modest gain in Fold 0 may stem from its simpler categories, which benefit less from richer descriptions.

To further validate the benefits of multi-text guidance, we compare it against single-text prompts on complex categories (Table~\ref{tab:text_comparison2}). Multi-text prompts consistently outperform their single-text counterparts, with notable mIoU gains for ``person" (+9.73\%) and “table” (+1.46\%), reflecting improved handling of semantic variability. Similarly, for structurally complex classes like “train” and “sofa,” gains of 6.53\% and 2.09\% indicate better separation from confusing backgrounds. These results confirm that multi-text descriptions enrich semantic priors and significantly enhance model generalization in challenging few-shot segmentation settings.

\subsubsection{Ablation study on text descriptions} 
We evaluate the impact of textual description quality and their combinations on the ``train" category using three descriptions (Table~\ref{tab:descriptions_comparison}): (1) a generic label (``Train"); (2) a moderately detailed phrase (``A train’s locomotive"); (3) a rich semantic description (``A train has a long and rectangular body with multiple connected cars”).
Using only Description 1 yields a baseline mIoU of 76.98\%. Replacing it with Description 2 or 3 increases performance to 82.93\% and 80.47\%, demonstrating that richer descriptions offer stronger semantic priors.
Next, we evaluate combinations of descriptions. Pairing Description 1 with 2 or 3 results in limited or negative effects (82.77\%, 80.68\%), likely due to semantic redundancy introduced by the generic prompt. In contrast, combining Descriptions 2 and 3 improves performance to 83.08\%, and using all three achieves the best result at 83.51\%. This shows that integrating diverse, informative prompts enhances segmentation by providing complementary semantic cues. The use of carefully designed multi-text prompts enables more comprehensive modeling of object characteristics, thereby validating the efficacy of our proposed multi-text strategy.

\begin{figure}[t]
	\centering
	\includegraphics[width=1\linewidth]{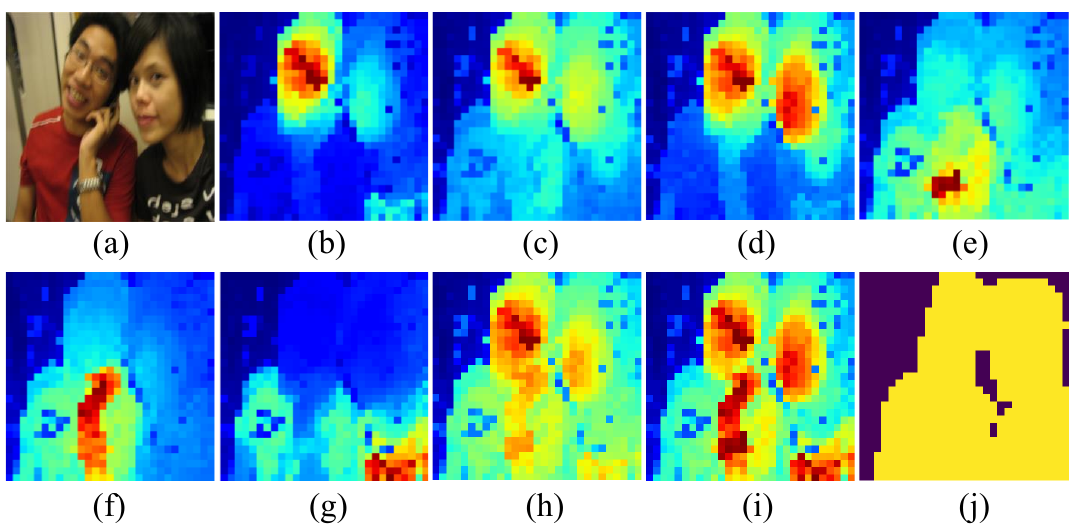}
	\caption{Visualization of refined text-guided priors generated from different textual descriptions and their aggregated forms.
                (a) Query image from the "person" category.
                (b)–(g) Individual priors generated using various textual descriptions:
                (b) “Person.”
                (c) “Person with clothes.”
                (d) “Person’s head.”
                (e) “Person’s arm.”
                (f) “Person’s hand.”
                (g) “Person with T-shirt.”.
                (h) Prior obtained by averaging the aggregation of the above priors.
                (i) Prior obtained by maximizing the aggregation of the above priors.
                (j) Ground-truth mask for the target object (person) in the image.
                }
	\label{fig:vistext2}
    \vspace{-2pt}
\end{figure}

Additionally, Fig.~\ref{fig:vistext2} illustrates how different descriptions activate distinct object parts in the "person" category. Aggregated priors via averaging or maximization offer broader and more semantically complete coverage, confirming the advantage of multi-text integration for robust segmentation guidance.

\begin{table}[t]
	\centering
	\caption{Comparison of different text descriptions and their combinations on the category ``train" for mIoU.}
	\label{tab:descriptions_comparison}
        \scalebox{0.85}{
	\begin{tabular}{ccccc}
		\toprule
		\textbf{Description 1} & \textbf{Description 2} & \textbf{Description 3} & \textbf{mIoU (\%)} \\
		\midrule
		\checkmark &  &  & 76.98 & \\
		& \checkmark &  & 82.93 &\\
		&  & \checkmark & 80.47 &\\
		\checkmark & \checkmark &  & 82.77 \\
		\checkmark &  & \checkmark & 80.68 \\
		&  \checkmark & \checkmark & 83.08 \\
		\checkmark & \checkmark & \checkmark & \textbf{83.51} \\
		\bottomrule
	\end{tabular}
        }
\end{table}

\begin{figure}[t]
	\centering
	\includegraphics[width=0.9\linewidth]{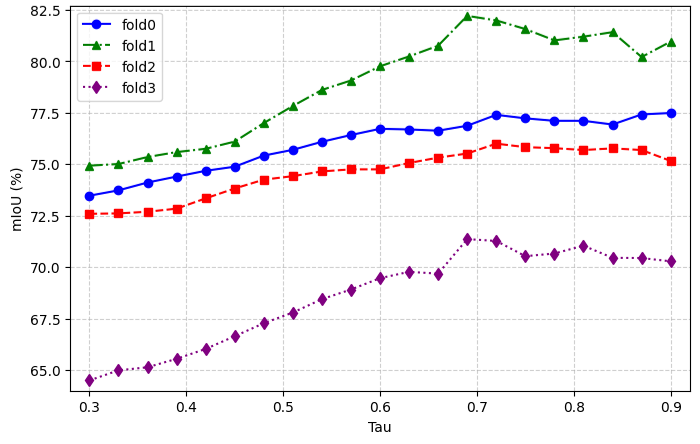}
	\caption{Impact of different thresholds in TSP on final results.}
	\label{fig:visths}
        \vspace{-2pt}
\end{figure}

\subsubsection{Ablation study on the threshold in TSP}

We evaluate the impact of the threshold $\tau$ by varying it from 0.3 to 0.9, with results across the four $\text{PASCAL-}5^i$ folds shown in Fig.~\ref{fig:visths}.

We observe a consistent trend across all folds. 
Lower thresholds ($\tau < 0.5$) enable broad propagation but introduce noise, especially harmful in complex scenes (e.g., Fold 3). Conversely, high thresholds ($\tau > 0.8$) restrict propagation too much, missing relevant regions and reducing performance—most notably in Folds 2 and 3.

These results confirm that both over- and under-propagation hinder segmentation. A moderate threshold provides a better trade-off between noise suppression and context retention. We thus select $\tau = 0.73$ as the optimal setting, offering consistent and robust performance across all folds.
\section{Conclusion}
We proposed MTGNet, a Multi-Text Guided Few-Shot Semantic Segmentation Network designed to address challenges such as incomplete foreground activation, high intra-class variations, and unreliable support features in FSS. Unlike prior methods that use a single prompt, MTGNet leverages multiple class-specific textual descriptions and employs a dual-branch architecture for deep cross-modal prior refinement.
The MTPR module aggregates diverse semantic cues for more complete activation, the TAFF module enhances support-query alignment using multi-text anchors, and the FCWA module boosts prior reliability by suppressing inconsistent regions via self-similarity analysis.
Extensive experiments on $\text{PASCAL-}5^i$ and $\text{COCO-}20^i$ demonstrate MTGNet’s effectiveness, with notable improvements on folds
exhibiting high intra-class variations.
In future work, we aim to incorporate adaptive, context-aware, and dynamic cross-modal reasoning to further enhance the generalization and robustness of semantic priors in open-world segmentation scenarios.

\section*{Acknowledgment}
This work is supported by the National Natural Science Foundation of China under Grant No.61803290 and No.61773301, and by the Fundamental Research Funds for the Central Universities under Grant No.ZYTS24022.     

{\small
\bibliographystyle{ieeetr}
\bibliography{egbib}
}

\vspace{-4em}
\begin{IEEEbiography}[{\includegraphics[width=1in,height=1.25in,clip,keepaspectratio]{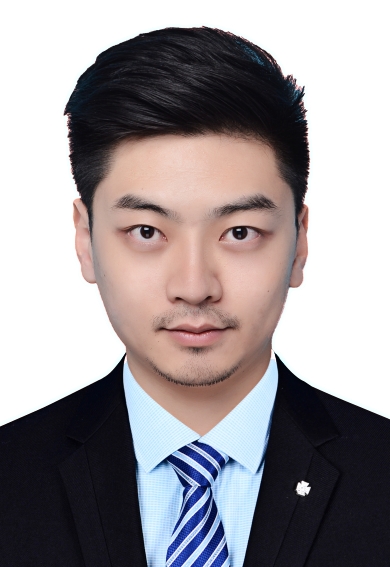}}]{Qiang Jiao} received the B.S. and Ph.D. degrees from Nanjing University of Science and Technology, Nanjing, China, in 2010 and 2017, respectively. He is currently with the School of Mechanoelectronic Engineering, Xidian University, Xi’an, China. His current research interests include computer vision and few-shot learning.
\end{IEEEbiography}
\vspace{-4em}
\begin{IEEEbiography}[{\includegraphics[width=1in,height=1.25in,clip,keepaspectratio]{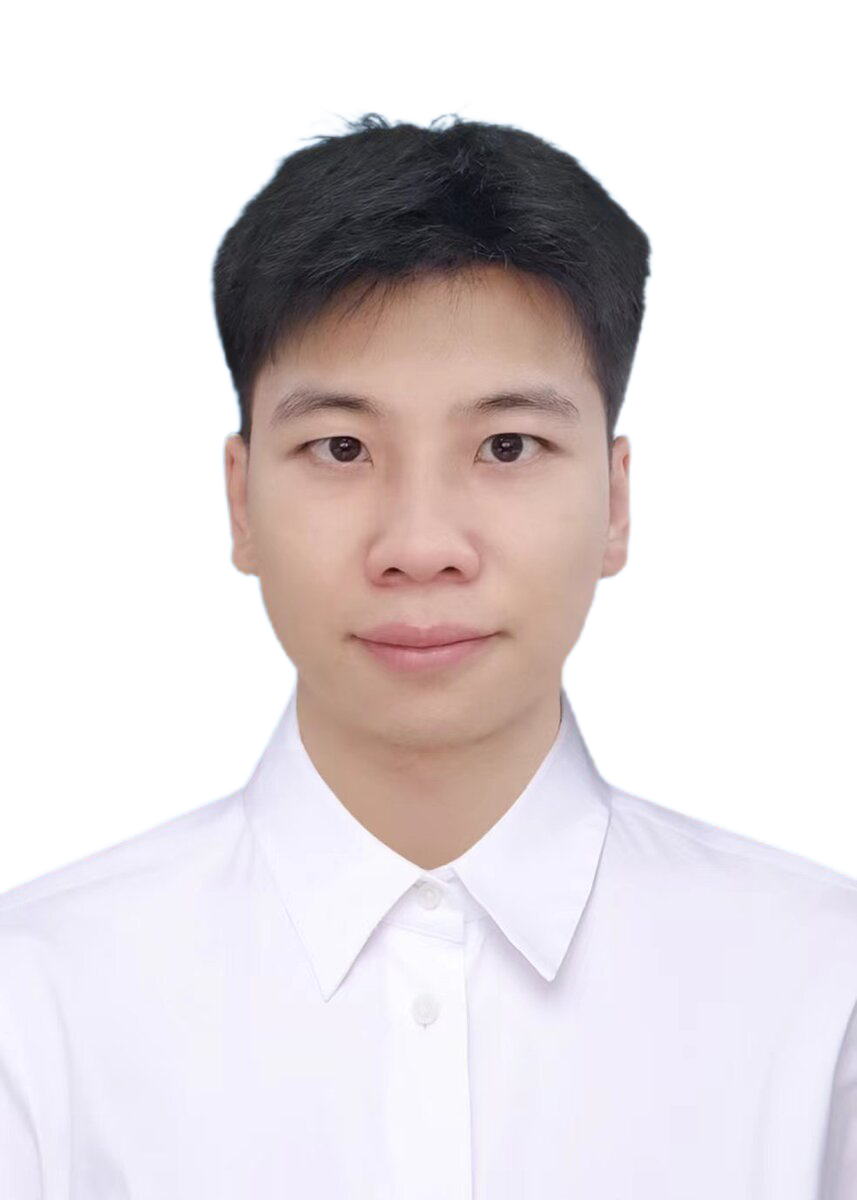}}]
{Yan Bin} received the B.E. degree in engineering from Wuhan University of Technology, Wuhan, China, in 2020.  
He is currently pursuing the master’s degree in the School of Mechano-Electronic Engineering, Xidian University, Xi’an, China.  
His research interests include computer vision and few-shot learning.
\end{IEEEbiography}
\vspace{-4em}
\begin{IEEEbiography}[{\includegraphics[width=1in,height=1.25in,clip,keepaspectratio]{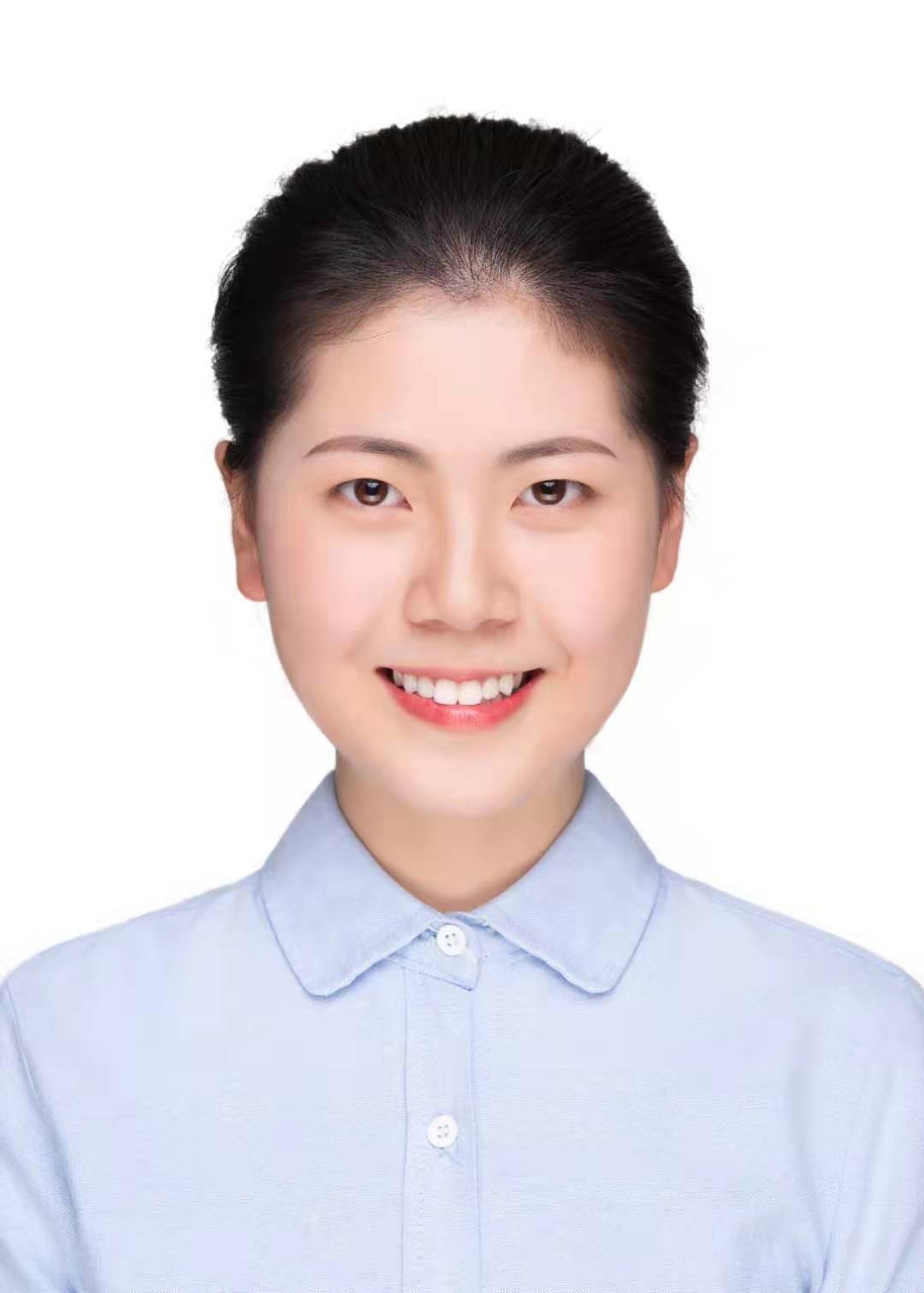}}]
{Yi Yang} received the B.E. degree in 2023 from Tianjin Normal University, Tianjin, China.  
She is currently pursuing the master’s degree in School of Mechano-Electronic Engineering, Xidian University, Xi’an, China.
Her research interests include computer vision and few-shot learning.
\end{IEEEbiography}
\vspace{-4em}
\begin{IEEEbiography}[{\includegraphics[width=1in,height=1.25in,clip,keepaspectratio]{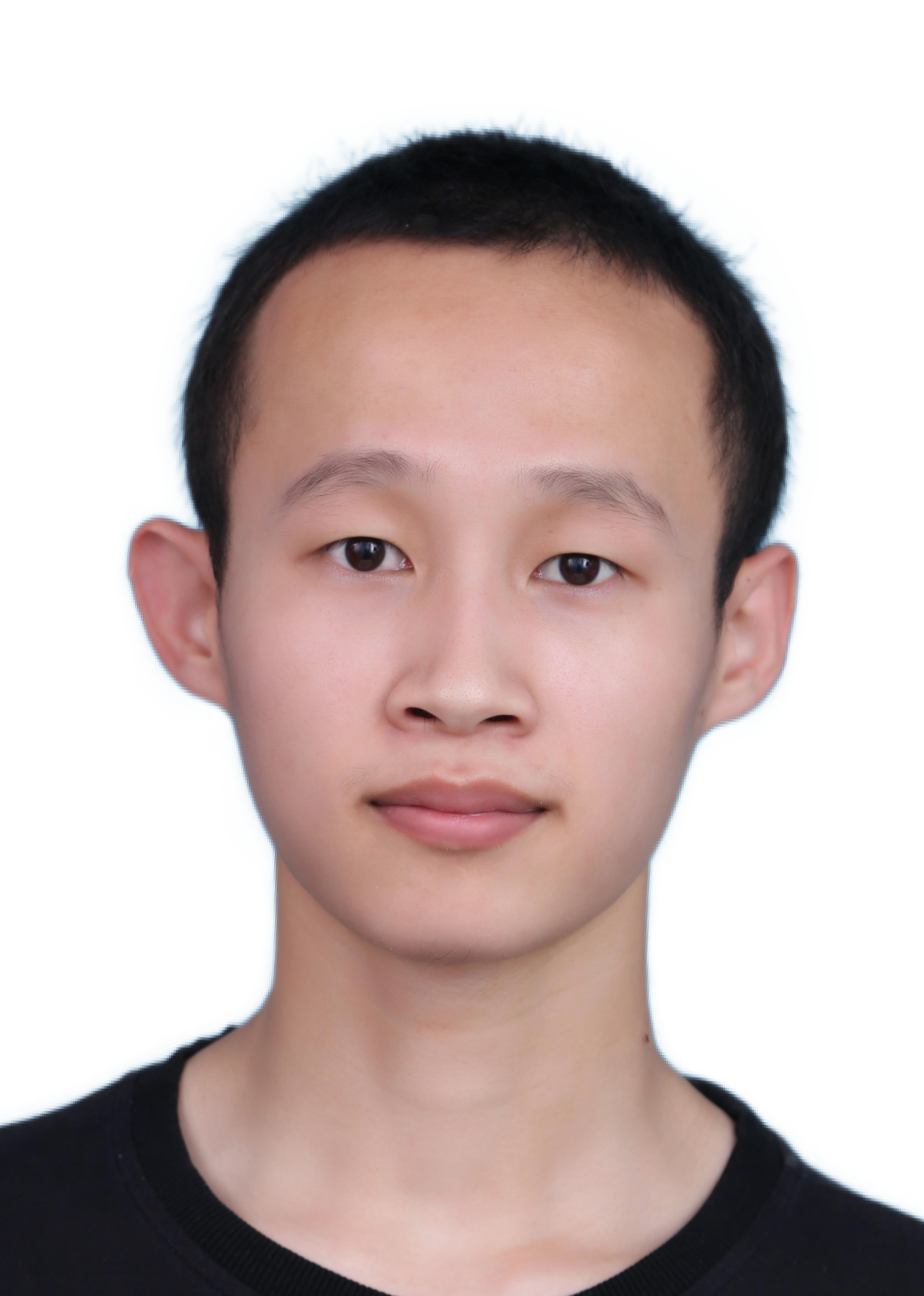}}]
{Mengrui Shi} received the B.E. degree in 2025 from Xidian University, Xi'an, China. 
He is currently pursuing the master’s degree in the School of Mechano-Electronic Engineering, Xidian University, Xi’an, China.  
His research interests include computer vision and few-shot learning.
\end{IEEEbiography}
\vspace{-4em}
\begin{IEEEbiography}[{\includegraphics[width=1in,height=1.25in,clip,keepaspectratio]{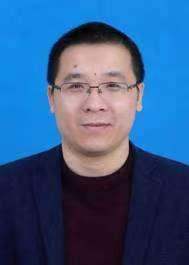}}]{Qiang Zhang} received the B.S. degree in automatic control, the M.S. degree in pattern recognition and intelligent systems, and the Ph.D. degree in circuit and system from Xidian University, China, in 2001, 2004, and 2008, respectively. He was a Visiting Scholar with the Center for Intelligent Machines, McGill University, Canada. He is currently a professor with the Automatic Control Department, Xidian University, China. His current research interests include computer vision and image processing.
\end{IEEEbiography}

\end{CJK}
\end{document}